\documentclass[journal]{IEEEtran}
\usepackage{color}
\usepackage{amsmath}
\usepackage[noend]{algpseudocode}
\usepackage{algorithmicx,algorithm}
\usepackage{graphicx}
\usepackage{caption} % DO NOT CHANGE THIS AND DO NOT ADD ANY OPTIONS TO IT
\usepackage{graphicx}
\usepackage{verbatim}
\usepackage{multirow}
\usepackage{bbding}
\usepackage{bbding}
\usepackage{pifont}
\usepackage{makecell}
\usepackage{booktabs} %
\usepackage{hyperref}
\usepackage{url}

\hypersetup{hidelinks} 

\usepackage[switch]{lineno}
\begin{document}
%
%\linenumbers 
\title{Duality-Gated Mutual Condition Network for RGBT Tracking}

\author{Andong Lu,~%~\IEEEmembership{Fellow,~OSA,}
		Cun Qian,
		Chenglong Li,~%~\IEEEmembership{Member,~IEEE,}
		Jin Tang,
      and Liang Wang
		
		% <-this % stops a space
	\thanks{
Manuscript received November 30, 2020; revised June 19, 2021 and November 23, 2021; accepted March 31, 2022.
This work was supported in part by the Major Project for New Generation of AI under Grant 2018AAA0100400;
in part by the National Natural Science Foundation of China under Grant 61976003, Grant 62076003, and Grant 61860206004;
in part by the Science and Technology Project of State Grid Anhui Electric Power Company Ltd. under Grant 5212S01900RN; 
and in part by the Open Project Program of the National Laboratory of Pattern Recognition (NLPR).(Corresponding author: Chenglong Li.).
%
%The authors are with the Key Laboratory of Intelligent Computing and Signal Processing, Ministry of Education, School of Computer Science and Technology, Anhui University, Hefei 230601, China, and also with the Anhui Provincial Key Laboratory of Multimodal Cognitive Computation, School of Computer Science and Technology, Anhui University, Hefei 230601, China
%
A. Lu, C. Qian and J. Tang are with Anhui Provincial Key Laboratory of Multimodal Cognitive Computation, School of Computer Science and Technology, Anhui University, Hefei 230601, China. 
C. Li is with Information Materials and Intelligent Sensing Laboratory of Anhui Province, Anhui Provincial Key Laboratory of Multimodal Cognitive Computation, School of Artificial Intelligence, Anhui University, Hefei 230601, China. 
L. Wang is with the National Laboratory of Pattern Recognition, Institute of Automation, Chinese Academy of Sciences, Beijing 100190, China.
(e-mail: adlu\_ah@foxmail.com; hongyuqian934@gmail.com; lcl1314@foxmail.com; tangjin@ahu.edu.cn; wangliang@nlpr.ia.ac.cn).
}% <-this % stops a space
	%\thanks{}% <-this % stops a space
}


\markboth{IEEE Transactions on Neural Networks and Learning Systems}%
{Shell \MakeLowercase{\textit{et al.}}: Bare Demo of IEEEtran.cls for IEEE Journals}



\maketitle

\begin{abstract}
Low-quality modalities contain not only a lot  of noisy information but also some discriminative features in RGBT tracking. However, the potentials of low-quality modalities are not well explored in existing RGBT tracking algorithms. 
In this work, we propose a novel duality-gated mutual condition network to fully exploit the discriminative information of all modalities while suppressing the effects of data noise.
In specific, we design a mutual condition module, which takes the discriminative information of a modality as the condition to guide feature learning of target appearance in another modality. 
Such module can effectively enhance target representations of all modalities even in the presence of low-quality modalities.
To improve the quality of conditions and further reduce data noise, we propose a duality-gated mechanism and integrate it into the mutual condition module.
To deal with the tracking failure caused by sudden camera motion, which often occurs in RGBT tracking, we design a resampling strategy based on optical flow.
It does not increase much computational cost since we perform optical flow calculation only when the model prediction is unreliable and then execute resampling when the sudden camera motion is detected.
Extensive experiments on four RGBT tracking benchmark datasets show that our method performs favorably against the state-of-the-art tracking algorithms.
\end{abstract}


% \fi
\IEEEpeerreviewmaketitle

\section{Introduction}
\IEEEPARstart{R}{GBT} tracking, a popular research stream of visual tracking, aims at estimating the states of target object in a RGBT sequence given the initial ground truth bounding box in the first frame pair. 
Benefiting from the strong complementary advantages of RGB and thermal infrared data, RGBT trackers could work well in all-day and all-weather conditions. 
%
%{\color {red}
%Similarly, more and more research fields~\cite{Li2021UAVHumanAL, Dou2020UnpairedMS, Liu2020LearningSS, Quan2021HolisticLF} exploit multi-modal information to handle tasks in open scenes, which shows the effectiveness of multi-modal information.}
%
Therefore, RGBT tracking receives more and more attentions and has achieved astonishing progress in recent years~\cite{Li17rgbt210,lan2018robust,wang2018learning,Li18eccv,zhu2019dense, zhang2019mfdimp, zhang2020MaCNet,2020CMPP,2020JMMAC,2020CAT}. 
%
%Benefit from the strong complementary advantages of the two modality features, RGBT tracking receives more and more attentions and has achieved astonishing progress in recent years~\cite{Li17rgbt210,xu2018relative,lan2018modality,lan2018robust,wang2018learning,Li18eccv,li2019manet, zhu2019dense, zhang2019multi, zhang2019siamft, zhang2020MaCNet}. 
%

%\begin{figure}[t]
%	\centering	\includegraphics[width=0.4\textwidth]{paper_fig/attibute_map_vv} % Reduce the figure size so that it is slightly narrower than the column.
%	\caption{Performance comparison of our DMCNet against several state-of-the-art methods on two typical subsets on RGBT234 dataset~\cite{li2019rgb}, including CAT~\cite{2020CAT}, CMPP~\cite{2020CMPP} and JMMAC~\cite{2020JMMAC}.
%	}
%	\label{fig::attribute_map}
%\end{figure}

%
Recent RGBT tracking methods mainly study how to effectively fuse RGB and thermal modalities.
One aspect is to introduce modality weights for adaptive fusion of different modalities.
For example, Zhang~\emph{et al.}~\cite{zhang2020MaCNet} propose modality-aware attention network to generate modality weights for adaptive fusion of different modalities with competitive learning.
Zhu~\emph{et al.}~\cite{2020FANet} propose a quality-aware feature aggregation network, which models each modality separately and then integrate different modality features by learning modality weights that represent qualities of different modalities.
%
%However, these methods usually have sub-optimal performance due to two major reasons.
%
%First, in a modality some cues are useful for tracking and others are not, and thus using modality weights only to judge contributions of different modalities sometimes is not reasonable.
%
%Second, data distributions of different modalities are not consistent, which might make the computation of modality weights be dominated by a single modality.
Another one aspect is to learn powerful features of each modality and then fuse them by ad-hoc ways.
For example, Zhang~\emph{et al.}~\cite{zhang2019mfdimp} propose a two-stream network structure and use a lager-scale generated RGBT dataset to train the network to learn the characteristics of each modality.
Li~\emph{et al.}~\cite{li2019manet} introduce three types of adapters to capture modality-specific, modality-shared and instance-aware target representations.

\begin{figure*}[t]
	\centering	
	\includegraphics[width=0.95\textwidth]{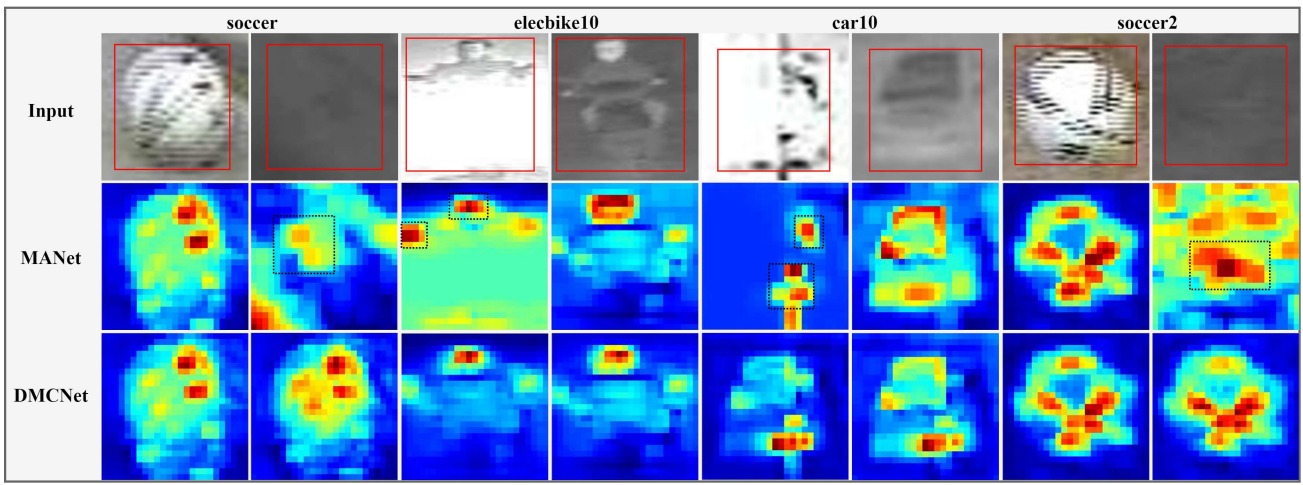} % Reduce the figure size so that it is slightly narrower than the column.
	\caption{Illustration of the effectiveness of our duality-gated mutual condition network (DMCNet) against a typical tracker MANet~\cite{li2019manet} on four examples in {\bf soccer}, {\bf elecbike10} , {\bf car10} and {\bf soccer2} respectively.
	Herein, the first row represents the input RGB and thermal frames, and the second row denotes the feature maps of the baseline method MANet, which does not include the duality-gated mutually conditional modules. The third row indicates the feature maps of our DMCNet. Here, the red rectangle box represents the tracking window, and the black dotted rectangle box indicates the discriminative information in low-quality modalities.}
	\label{fig::feature_maps}
\end{figure*}

However, all these methods do not explore the potentials of low-quality modalities well, which play a critical role in boosting RGBT tracking performance. An example is shown in Fig.~\ref{fig::feature_maps}.
In the first row, one can see that each RGBT image pair contains a lot of noisy information and the quality of one modality in each pair is extremely low.
%
%Existing RGBT tracking methods such as MANet~\cite{li2019manet}, the features learned from these modallity data will be as shown in the middle row of Fig.~\ref{fig::feature_maps}
%
The feature maps generated by a typical method MANet~\cite{li2019manet} are shown in the second row, and we can find that these features suffer from the effects of a lot of noises especially in the low-quality modalities.
These noise effects would degrade tracking accuracy and robustness.
%
%Therefore, how to better leverage different qualities of modal information to achieve effective interactions of modal features, which tend to plays a critical role in feature enhancement and noise reduction.
%
In addition, we observe that low-quality modalities usually contain some discriminative features which are useful for target localization, as shown in the second row of Fig.~\ref{fig::feature_maps}.
Therefore, simple suppression or removal of low-quality modalities can not fully explore the potentials of multi-source data.
To handle these issues, we propose a novel Duality-gated Mutual Condition Network (DMCNet) to fully exploit the discriminative information of all modalities while suppressing the effects of data noises.
In real-world scenarios, some modalities are sometimes unreliable due to the existence of adverse environments like total darkness, bad weathers and thermal crossover.
To make the full use of potentials of these modalities while suppressing the effects of data noises, we design a mutual condition module, which takes the discriminative information of a modality as the condition to guide feature learning of target appearance in another modality.
Moreover, we enhance feature representations of target appearance through a multi-scale convolutional layer and integrate it into the mutual condition module. 
Because we use features of one modality as the condition of the other modality, and some noises are thus inevitably included in conditions. Meanwhile, the features guided by conditions also contain noisy information.
To improve the quality of conditions and further reduce feature noises, we propose a duality-gated mechanism.
Fig.~\ref{fig::feature_maps} presents some examples to verify the effectiveness of the proposed duality-gated mutual condition network.
As shown in the third row, noises in modalities (especially for low-quality modalities) are significantly suppressed and discriminative abilities of target features are greatly boosted.

In addition, we find that the tracking performance is easily affected by the challenge of sudden camera motion, which frequently occurs in RGBT tracking task.
The major reason is that under such challenge search windows are hardly cover target objects, which would lead to tracking failure.
Common attempts are to expand search region~\cite{siamrpn_2018} and perform global search~\cite{zhu2016beyond}, but these methods bring more background information and thus increase the risk of model drift.
Meanwhile, the computational cost is usually greatly increased.
To deal with this problem, we develop a simple yet effective resampling method based on a fast optical flow algorithm, DisFlow~\cite{disflow}.
With a predefined threshold, we can judge whether the sudden camera motion occurs or not.
If occurs, we resample candidate target regions along the direction and magnitude of camera motion.
%The opposite direction of camera motion
%
Note that our resampling method does not increase computational cost much since we execute it only when the tracking failure caused by sudden camera motion is detected and the optical flow computation is only performed on the local regions around target objects.
The major contributions of this paper are summarized as follows. 
\begin{itemize}
\item We propose an effective approach to handle low-quality modalities in RGBT tracking. The approach is able to enhance the discriminative ability and suppress the effects of data noises of low-quality modalities and thus achieves large improvements in tracking accuracy and robustness.
\item We design a duality-gated mutual condition module to bi-directionally take the discriminative information of a modality as the condition to adaptively guide feature learning of target appearance in another modality.
\item We develop a simple yet effective resampling mechanism to deal with the tracking failure caused by the challenge of sudden camera motion, with a modest impact on tracking speed.
\item Extensive experiments on four RGBT tracking benchmark datasets, including GTOT~\cite{li2016learning}, RGBT210~\cite{Li17rgbt210}, VOT2019-RGBT~\cite{vot-rgbt2019} and RGBT234~\cite{li2019rgb} are conducted.
The results show that our tracking approach achieves the outstanding performance comparing with the state-of-the-art methods.

\end{itemize}

\begin{figure*}[!htb]
	\centering	\includegraphics[width=0.95\textwidth]{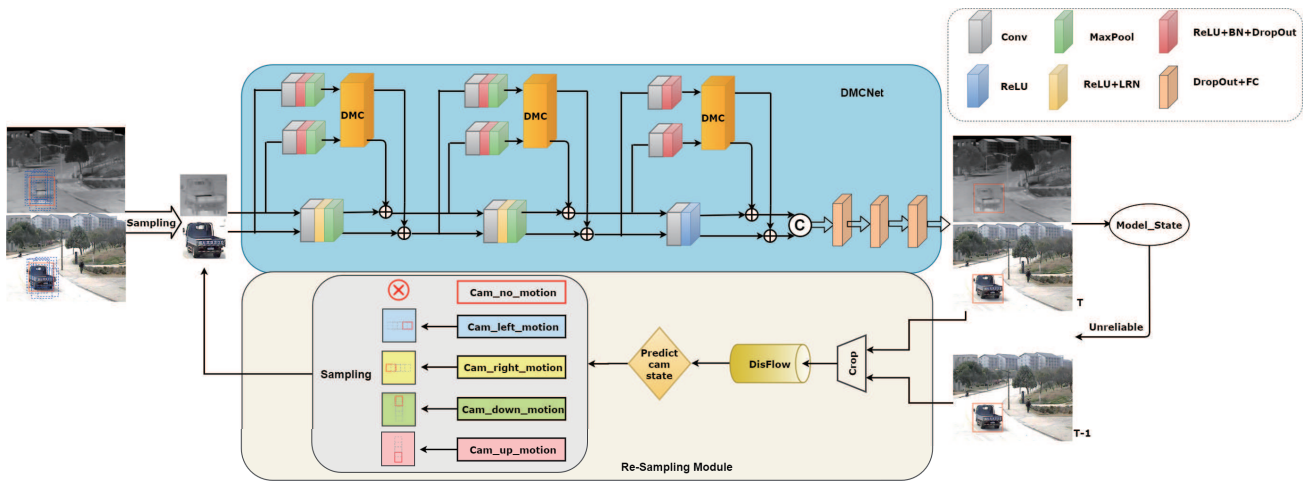} % Reduce the figure size so that it is slightly narrower than the column.
	\caption{Overall network architecture of our DMCNet. 
		Herein, $\oplus$, $\copyright$ denote the operations of element-wise addition and concatenation respectively. 
	}
	\label{fig::network_structure}
\end{figure*}

%I wish you the best of success.

 

\section{Related Work}

\subsection{RGBT Tracking Methods}
In recent years, more and more RGBT tracking algorithms has proposed~\cite{li2016learning,Li17tsmcs,Li17rgbt210,lan2018robust,li2018fusing,Li18eccv,zhang2019mfdimp,2020CAT,2020CMPP,2020JMMAC,2021MANet++}, making this field remarkable development.
Recent works~\cite{Li17tsmcs,li2016learning,lan2018robust} propose to learn modality weights to guide adaptive fusion of RGB and thermal modalities via reconstruction residues or classification scores.
However, these methods are susceptible to interference from low-quality modal information, and unreliable reconstruction residues or classification scores would lead to inaccurate weight computation.
Some recent studies~\cite{Li17rgbt210,li2018fusing,Li18eccv} are focus on learning robust feature representations of RGB and thermal modalities.
A graph learning approach is proposed by Li \emph{et al.}~\cite{Li17rgbt210} that constructs a patch-based weighted RGBT feature descriptor, and performs online tracking using the structured SVM. 
Several improvements~\cite{Li17rgbt210,li2018fusing,Li18eccv} are made in this research stream.
However, these methods are based on handcrafted features and their performance is easily affected by challenging factors.
Zhu~\emph{et al.}~\cite{2020FANet} propose a quality-aware feature aggregation network to integrate different modal features by learning modality weights that represent qualities of different modalities. 
Zhang et al.~\cite{zhang2019mfdimp} use different levels of fusion strategies to integrate the information of RGB and thermal modalities adaptively in an end-to-end deep learning framework. 
Wang et al.~\cite{2020CMPP} propose a cross-modal pattern-propagation tracking method to model intra-modal paired pattern-affinities, which reveal the latent cues between heterogeneous modalities.
Li et al.~\cite{2020CAT} propose a challenge-aware network to model the representation of modality-shared and modality-specific challenges.
Zhang et al.~\cite{2020JMMAC} employ a later fusion network that combines with motion tracker to jointly model appearance and motion cues for RGBT tracking.
Lu et al.~\cite{li2019manet,2021MANet++} propose a multi-adapter network to jointly
perform modality-shared, modality-specific and instance-aware target representation learning for RGBT tracking.
However, these methods do not explore the potentials well of low-quality modalities, which play a critical role in feature enhancement and noise reduction.

\subsection{Deep Mutual Learning}
The deep mutual learning aims to learn an ensemble of students collaboratively and teach each other using the losses based on the Kullback Leibler (KL) Divergence.
Zhang et al.~\cite{Zhang2017Deep} propose a mutual learning framework which composes of two untrained student networks, and use two mimicry loss to guide learning of student networks.
Yang et al.~\cite{2020MutualNet} propose the width-resolution mutual learning method to train a cohort of sub-networks with different widths using different input resolutions with KL loss.
Wu et al.~\cite{2020A} propose a mutual learning module, which composes three student branches, and use a L2-based mimicry loss to optimize the network.
Dou et al.~\cite{Dou2020UnpairedMS} design a loss inspired by knowledge distillation, which is used to distill semantic knowledge from high-resolution feature maps before softmax.
Liu et al.~\cite{Liu2020LearningSS} propose the selective self-mutual attention module to propagate long range contextual dependencies, and thus incorporate multi-modal information to learn attention and propagate contexts more accurately.
Li et al.~\cite{Li2021UAVHumanAL} propose a guided transformer network, which uses RGB videos as the guidance information, and constrains the learning of the transformers by applying the Kullback-Leibler divergence loss between the RGB features and transformed features.
These methods usually use the mimicry loss to supervise multi-branch networks collaboratively, where different initializations are used in different branches and the overall performance is thus improved by the mutual learning. 
Instead of mutual supervision by mimicry loss in existing works, our duality-gated mutual condition network is to exploit discriminative information of modalities as mutual conditions to enhance target representations of all modalities while mitigating noise effects.
%\IEEEpubidadjcol

\section{Duality-Gated Mutual Condition Network}
In this section, we first overview the our backbone architecture, and then introduce the details of the duality-gated mutual condition module and the resampling mechanism.
Fig.~\ref{fig::network_structure} shows the overall framework of our tracking method, which consists of the backbone network, the duality-gated mutual condition module and the resampling module.
In the following, we present the details of each part.

\subsection{Backbone architecture}
As shown in Fig.~\ref{fig::network_structure}, the input of the backbone network is the candidate patches cropped from aligned RGB and thermal image pair, and these patches are resized to the size of 107$\times$107.
In specific, we use sampling methods to generate a set of anchor boxes during the training and inference. 
For each anchor box, we crop a RGB patch and a thermal patch and take these two patches as the candidate patch pair which is inputted into our network to compute its prediction score.
Note that we use different sampling methods in the training phase and inference phase, and we will describe the details of these sampling methods in the following sections.
%{\color{red}
%To crop candidate patches from aligned RGB and thermal image pairs, we divide this process %into two steps. 
%
%The first step is to generate a series of anchor boxes, note that 
%
%The second step is to directly use these anchor boxes to crop the image pairs because two modalities share these anchor boxes. 
%
%}
%
%Our baseline algorithms of our network is MDNet~\cite{li2019manet}, which treats 
%
Our backbone is borrowed from the first three convolutional layers of VGG-M~\cite{simonyan2014very}, and their convolutional kernel sizes are 7$\times$7$\times$96, 5$\times$5$\times$256, 3$\times$3$\times$512 respectively.
The first and second convolutional layers are followed by a ReLU activation function, a local response normalization (LRN) and a max pooling layer.
The third convolutional layer is just followed by a ReLU activation function.
Followed by the last convolutional layer, the binary classification is performed, which consists of three fully connected layers with the output dimensions as 512, 512 and 2 respectively. 
And then we employ the multi-domain learning strategy to model appearance variations of instance objects~\cite{nam2016learning}.
As in MANet~\cite{li2019manet} to model target representations robustly using multi-modal information, we use the modality adapters to extract modality-specific features.
In specific, the modality adapters are composed of convolutional layers, a ReLU activation function, a batch normalization, a dropout and a max pooling layer, and the settings of modality adapters in different modalities are same. 
The sizes of the convolutional kernels are 3$\times$3$\times$96, 1$\times$1$\times$256, 1$\times$1$\times$512 in the three levels, as shown in Fig.~\ref{fig::network_structure}.
Therefore, we employ the backbone network to extract modality-shared features, and model modality-specific features using the modality adapters, i.e modality-specific sub-network.

\subsection{Duality-Gated Mutual Condition Module}
Although above backbone can provide robust target representations, the interactions between modalities are ignored, which play a critical role in strengthening discriminative ability of multi-modal representations while suppressing feature noises.
During the training or inference phase, we do not know which modality is better or worse and the accurate estimation of modality quality is very difficult. 
Therefore, we design a mutual feature modulation module to adaptively learn effective features from all modalities.	
Herein, the mutual feature modulation contains RGB-to-T and T-to-RGB feature modulations, which are executed simultaneously in a bi-directional manner and make noises from both modalities be alleviated by the designed duality-gated scheme.
	
%To handle this problem, we propose a novel duality-gated mutually condition module with bi-directional parallel architecture, which achieves bi-directional conditional feature modulation between two modalities.

%Since, In real scenarios, we do not know which modality is reliable in training and testing. 
%

{\flushleft \bf RGB-to-T Feature Modulation}.
%
%Inspired by FILM~\cite{perez2018film}, many works~\cite{Wang2018RecoveringRT,2020CAT,Li2020SegmentingOI} employ feature-wise affine transformation 
%
Feature modulation is an effective method to influence or change the output feature of a model.
We want to leverage discriminative features of RGB data to guide feature learning of target appearance in thermal data, and thus design a scheme to modulate thermal features with the RGB information as conditions.
Our idea is inspired by FiLM~\cite{perez2018film}, which use prior information to construct two conditions that scale and shift features respectively.
However, some issues need to be addressed when we apply FiLM to our task.
First, the diversity of scaling and shifting conditions is low and the potential of conditional feature learning could not be fully explored.
Second, RGB information might contain noises since we do not know whether it is high-quality or not, and RGB-based conditions might thus be harmful for feature learning of thermal data.
   
%
%In specific, we firstly modulate output of thermal modality adapter by scaling with the features of RGB modality as the condition.
%
To handle the first issue, we propose a new scheme to generate high-quality and diverse scaling and shifting conditions.
We first modulate the outputs of thermal modality adapter by applying a multi-scale scaling transformation, based on the outputs of RGB modality adapter.
In specific, we design a $MSConv$ layer (denoted as $\emph{W}^{ms}$), which is implemented differently in different layers, to capture multi-scale feature information for generating the scaling conditions. 
In the first layer, the receptive fields of feature maps are small and the inter-modal variations are large, and we thus use four different convolutions to capture multi-scale information from different sizes of receptive fields.
We use the $1\times1$ and $3\times3$ convolutions to capture local details, and use the $3\times3$  dilated convolution with the dilated rate of 2 and the $5\times5$ convolution to model global information.
In the middle and high layers, the $MSConv$ layer is implemented with a combination of $1\times1$, $3\times3$, and $1\times1$ convolutions, respectively.
The multi-scale scaling conditions of RGB modality can be expressed as follows: 
\begin{equation}
	\begin{aligned}
		&f^{ms}_{R} = ({\emph{W}^{f}} \ast [ {\emph f}_{R} \ast \emph{W}^{ms} ])\\ 
		%	&f^{ms}_{T} = ({\emph{W}^{f}} \ast [ {\emph f}_{T} \ast \emph{W}^{ms} ]) 
	\end{aligned}
	\label{eq::1}
\end{equation}
where $f_{R}$ is the feature maps of the modality adapter in RGB modality, $\ast$ represents the convolutional operation, and $f^{ms}_{R}$ denotes the multi-scale scaling conditions. 

\begin{figure*}[h]
	\centering	\includegraphics[width=0.95\textwidth]{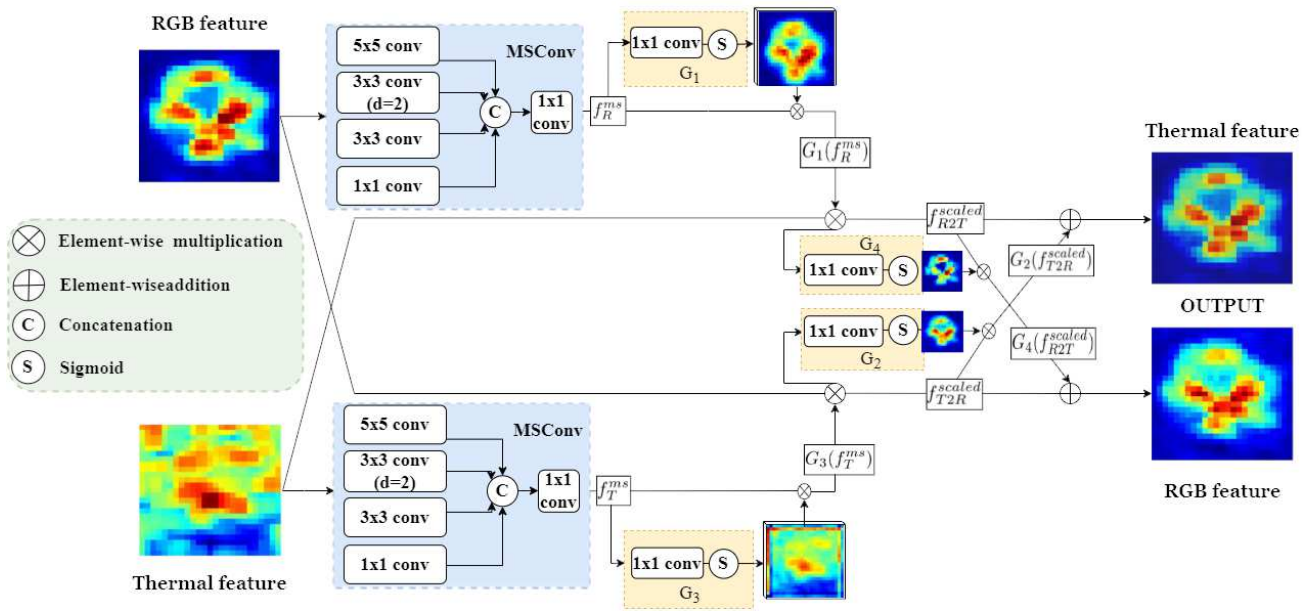} % Reduce the figure size so that it is slightly narrower than the column.
	\caption{Illustration of our duality-gated mutual conditional module, in which we take feature maps of $soccer2$ as an example.  	
	}
	\label{fig::sub_network}
\end{figure*}

Then, we further modulate the above scaled features of thermal modality adapter by applying a multi-modal shifting transformation, based on both outputs of RGB and thermal modality adapters.
On one hand, we generate the multi-scale features from thermal modality since these features will be beneficial to enhancing target representations in thermal modality.
On the other hand, we fuse multi-scale thermal features and RGB features to form the high-quality rich shifting conditions.
The details of condition generation are shown in Fig.~\ref{fig::sub_network}.   
Thus we can express the RGB-to-T feature modulation as follows:
\begin{equation}
	\begin{aligned}
		&f^{out}_{T}= f_T \odot f^{ms}_{R} +  f^{scaled}_{T2R} \\
	\end{aligned}
	\label{eq::2}
\end{equation}
where $\odot$ represents the elemental-wise multiplication, and $f^{out}_{T}$ denote the modulated thermal features.
%used in the RGB-to-T modulation, but it is
Note that $f^{scaled}_{T2R}$ is generated in the T-to-RGB modulation.
In specific, $f^{scaled}_{T2R}$ represents the feature ($f_R$) of RGB modality modulated by the multi-scale feature ($f_T^{ms}$) of thermal modality. 
Moreover, we employ a gate ($G_3$) to improve the quality of multi-scale feature as shown in the Fig~\ref{fig::sub_network}.
Therefore, we formulate it as follows:
\begin{equation}
	\begin{aligned}
		&f^{scaled}_{T2R} = f_R \odot G_3(f_T^{ms}) \\
	\end{aligned}
	\label{eq::3}
\end{equation}
Similar to the computation of $f^{scaled}_{T2R}$, we can compute $f^{scaled}_{R2T}$, which is generated in the  RGB-to-T modulation and used in the T-to-RGB modulation.
%the shifting condition in the
\begin{figure}[t]
	\centering
	\includegraphics[width=0.45\textwidth]{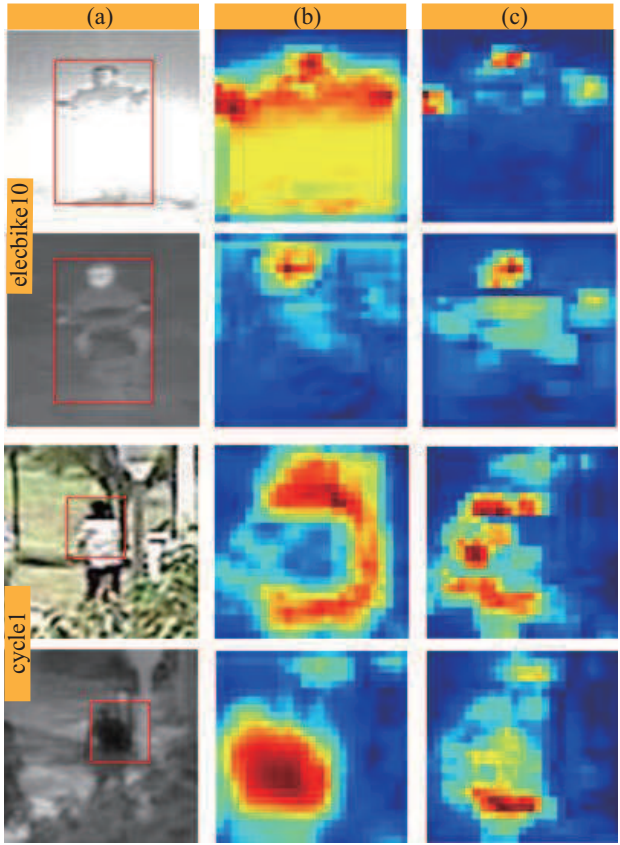} 
	\caption{Illustration of the effectiveness of our duality-gated scheme on two example frames. 
	Herein, the column (a) represents the input images of RGB and thermal modalities, and the column (b) denotes the response maps of $G_1$. The column (c) indicates the response maps of  $G_2$.
	}
	\label{fig::gate_maps}
\end{figure}

For the second issue, the noises from unreliable modalities might harm the quality of conditions, which would lead to the risk of feature degradation in learning.
To alleviate this issue, some works~\cite{Quan2021HolisticLF,Shahroudy2016NTURA} leverage LSTM, RNN and their variants to adaptively fuse multi-modal information.
For instance, Quan et al.~\cite{Quan2021HolisticLF} propose a holistic LSTM to incorporate information from pedestrians and vehicles adaptively.
There are also some works ~\cite{huang2021unsupervised, huang2019sbsgan} that try to suppress background noise through adversarial generative strategies.
However, these methods increase the memory cost owing to the application of memory cells.
Therefore, we design a duality-gated strategy without extra memory cost to avoid noisy information of RGB modality in generating conditions.
Fig.~\ref{fig::sub_network} shows the details of the duality-gated structure, where the two gates have the same internal structure, which is formulated as follows.
\begin{equation}
	\begin{aligned}
		&G =  \sigma(Conv(f))\\
	\end{aligned}
	\label{eq::4}
\end{equation}
where $Conv(\cdot)$ and $\sigma$ denote the operations of $1\times1$ convolution and sigmoid function respectively. $f$ indicates the input features.
Therefore, we embed the duality-gated formulation in RGB-to-T modulation as follow:
\begin{equation}
	\begin{aligned}
		&f^{out}_{T}= f_T \odot G_{1}(f^{ms}_{R}) +  G_{2}(f^{scaled}_{T2R}) \\
	\end{aligned}
	\label{eq::5}
\end{equation}
where $G_{1}$ and $G_{2}$ represent two gates to mitigate the noises of multi-scale scaling conditions and the fused features respectively.
The effectiveness of our duality-gated mechanism is shown in Fig.~\ref{fig::gate_maps}, and we can see that the designed two kinds of gate mechanisms can effectively avoid the noises of single modality and multiple modalities in information propagation.
Here, the column (b) shows that the noises of a single modality features are suppressed and some discriminative features are enhanced by $G_1$.
For example, in the $elecbike10$ sequence, the feature response of glare region in RGB modality is enhanced by $G_1$.
In the $cycle1$ sequence, $G_1$ well suppresses the response of non-target region in candidate patch. 
The column (c) demonstrates that $G_2$ effectively filters noise propagation of multi-modal features.
For instance, in the $elecbike10$ sequence, the glare information of RGB modality is regarded as noisy information, and thus it is well suppressed by the $G_2$ in the propagation to thermal modality.
Moreover, the body of target information in thermal modality is useful for RGB modality, which is effectively enhanced by $G_2$. 
%es
Therefore, the duality-gated mechanism can well suppress the noises in both single-modal features and multi-modal features.

{\flushleft \bf T-to-RGB Feature Modulation}.
In this work, we want to leverage all discriminative information of different modalities regardless of low-quality and high-quality modalities.
Therefore, we adopt a bi-directional conditional feature learning structure to fully mine the discriminative features of all modalities.
The structure of T-to-RGB feature modulation is symmetric to RGB-to-T one, and we thus obtain the final output of T-to-RGB feature modulation as follow:
\begin{equation}
	\begin{aligned}
		&f^{out}_{R}= f_R \odot G_{3}(f^{ms}_{T}) +  G_{4}(f^{scaled}_{R2T}) \\
	\end{aligned}
	\label{eq::6}
\end{equation}
where the $f^{out}_{R}$ represents the modulated RGB features.
%
%The general idea of the formula is that we employ RGB(Thermal) modality feature maps to scale the Thermal(RGB) modality feature maps by element-wise multiplication, which help to obtain a more favorable representation of the Thermal(RGB) modality under potentially modality space.
%
%When the distance of two potentially modality spaces become narrow due to the scaling process, it is more efficient to use the shift factor, and in the following experiment result also prove it.

%and then use shift factor feature will more effectivly when two feature space of modality is narrow by scaling process.

{\flushleft \bf Analysis of Duality-Gated Effects}.
Herein, we discuss how to avoid noisy information in generating conditions by the designed gates.
We take the RGB modality as example. 
(1) In RGB-to-T modulation, we generate scaling and shifting conditions using RGB modality. 
For the scaling condition, we use multi-scale features of RGB modality for generation.
Therefore, we use $G_1$ to filter noisy information in the generation of scaling condition.
For the shifting condition, we use the input features of RGB modality and multi-scale features of thermal modality for the generation, in which the noisy information of multi-scale thermal features are already filtered by $G_3$. 
Therefore, we utilize the $G_2$ to suppress noisy information of RGB modality in the generation of shifting condition . 
(2) In T-to-RGB modulation, we generate shifting condition using RGB modality and the scaling condition is only based on thermal modality. 
For the shifting condition, we use the multi-scale feature of RGB modality and the input thermal feature for generation. While the noisy information of multi-scale feature of RGB modality is already filtered by $G_1$. 
In summary, we can avoid noisy information of RGB modality in generating all conditions. 
The noisy information of thermal modality in generating conditions is filtered by our gates in a similar way.

\subsection{Re-Sampling Module}
\begin{figure}[t]
	\centering	\includegraphics[width=0.45\textwidth]{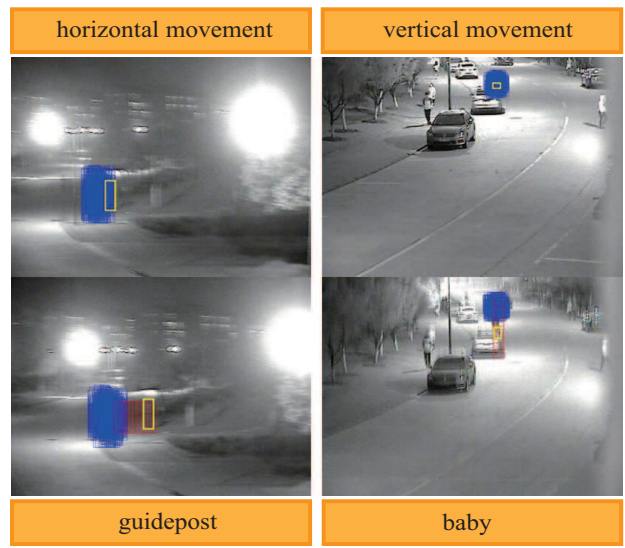} % Reduce the figure size so that it is slightly narrower than the column.
	\caption{Example of the re-sampling scheme. 
		Herein, the blue boxes represent the original sampling regions, the red boxes represent the re-sampling regions, and the yellow boxes are the ground truths. 
		Note that the images in first column involve horizontal movement of camera and the images in second column involve vertical movement of camera.
	}
	\label{fig::guide_samples}
\end{figure}

In RGBT tracking task, abrupt camera motion is a common challenge, which affects the performance much.
The major reason is that under such challenge the search window usually can not cover target objects, which would lead to tracking failure. 
Fig.~\ref{fig::guide_samples} shows the limitation of the widely used Gaussian sampling strategy.
When target has large displacement between two adjacent frames due to camera motion, the samples drawn by Gaussian sampling cannot cover the target since the displacement is beyond the searching window.
Common attempts are that one can expand search region~\cite{siamrpn_2018} and perform global search~\cite{zhu2016beyond}, but these methods bring more background information and thus increase the risk of model drift.
Meanwhile, the computational cost is usually greatly increased.

To handle these issues, we develop a re-sampling scheme based on a fast optical flow algorithm~\cite{disflow}, which can well cover the target with the re-sampling scheme guided by camera motion estimation.
As shown in Fig.~\ref{fig::guide_samples}, we can see that the anchor boxes (red box) generated by our resampling can cover the target region (yellow box) well, while anchor boxes (blue box) generated by Gaussian sampling cannot cover the target.
%
%Specifically, when model state is unreliable, we employ the optical flow to detect the motion state of camera, and then determine whether the re-sampling is executed or not.
%
%First, we start optical flow estimation when tracking failure is detected, i.e., the predicted target score is below 0.
We describe the resampling process in the following six steps.
First, we start optical flow estimation when the model state is unreliable.
%tracking failure is detected. 
%
We judge the model state based on the prediction score of the tracking model, since the score can reflect the confidence of candidate patches belonging to the target. 
%belong? belongs?
In specific, if the score is lower than 0, we take the state of the model as unreliable. The similar setting is used in MDNet~\cite{nam2016learning}.
Second, we use Disflow~\cite{disflow} to compute displacements of all pixels in a local region around target object, and then calculate the mean displacement vector $[dx,dy]$.
In this work, this local region is centered at the target position in previous frame and its size is three times of the size of target bounding box.
Third, we judge whether abrupt camera motion occurs or not by comparing the amplitude of $[dx,dy]$ with a predefined threshold $u$.
Here $|dx|$ and $|dy|$ represent the mean values of the horizontal displacement and the vertical displacement respectively.
If $|dx|$ or $|dy|$ is below $u$, we think the failure is not caused by abrupt camera motion and do not execute re-sampling. 
Otherwise, when the average value of the horizontal or vertical displacement is greater than the predefined threshold $u$, we think the camera has horizontal or vertical movement. 
%
%Specifically, the camera is moving horizontally when the $|dx|$ is greater than the  $u$, and the camera is moving vertically when the $|dy|$ is greater than the $u$.
%
Moreover, we judge the direction of camera motion (left/right or up/down) according to the sign of $|dx|$ and $|dy|$.
Fourth, we treat the opposite direction of the camera movement as the direction of target movement, which is performing re-sampling according to the camera motion state.
Fifth, we use the sampling range and direction of target movement to draw a set of candidates. 
In specific, we empirically re-sample 16 candidate regions along the target direction which is opposite to camera motion and the step of re-sampling is set as follows.
%
%Fourth, we perform re-sampling according to $[dx,dy]$.
%
%In specific, we empirically re-sample 16 candidate regions along the judged direction which is opposite to camera motion and the step of re-sampling is set as follows.
%
In the horizontal direction, we take the quarter of the width of target bounding box as the step, while in the vertical direction we take the quarter of the height of target bounding box.
Finally, we feed these samples into our network to compute their scores and combine them with the results of Gaussian sampling to compute the final predicted result.
%
%
%Such strategy can improve the robustness of re-sampling as the re-sampled candidates are sometimes unreliable.
%
We present the more details in Algorithm~\ref{alg::tracking_alg}.

\subsection{Network Training}

Our backbone network is MDNet~\cite{nam2016learning}, which is an RGB tracker and trained by a multi-domain learning algorithm on ImageNet dataset~\cite{imagenet_cvpr09}. 
Therefore, we load the above model parameters as the pre-trained parameters of our backbone network. 
The modality-specific sub-network is a series of small convolution layers which are parallel to the backbone network. 
Here, we randomly initialize these parameters, which is similar to MANet~\cite{li2019manet}.

We employ the Stochastic Gradient Descent (SGD) algorithm to train our network effectively.
%
%{\color{red}
%First, we initialize our backbone network using the above pre-trained model, and randomly initialize the modality-specific sub-network and duality-gated mutual condition sub-network. 
%
%Here, the backbone network contains three convolution layers and the first two fully connected layers.
%}
%
%First, we load the pre-trained model, which is trained on the ImageNet~\cite{deng2009imagenet} dataset, and then use a multi-domain learning algorithm~\cite{nam2016learning} to learn the parameters of the backbone sub-network and the first two fully connected layers, while the parameters of other sub-networks are initialized randomly.
%
In specific, we use RGBT dataset to train the whole network with 200 epochs by the softmax cross-entropy loss.
It is worth noting that we set different learning rates for different sub-networks.
The backbone sub-network is loaded with the pre-trained model, which has powerful capability in feature representation. 
To maintain the capability of the pre-trained model, we set a small learning rate for backbone network. 
While parameters of the modality-specific sub-network and duality-gated mutual condition sub-network are initialized randomly. Therefore, we set a larger one for fast convergence.
In specific, we set the learning rates of backbone sub-network, first two fully connected layers and binary classification layer to 0.001, and modality-specific sub-network and mutual-conditional sub-network to 0.002.
In each iteration, we select 8 frames which are randomly chosen in each video sequence from training dataset to construct a mini-batch.
Next, we employ a uniform distribution strategy to generate a series of anchor boxes, which are around the ground truth box in each frame.
Then we select the boxes whose IoU with ground truth box are larger than 0.7 as positive samples and the boxes whose IoU are smaller 0.5 as negative samples. 
Finally, we crop 32 positive samples and 96 negative samples according to the above positive and negative anchor boxes in each frame to form the input data in a mini-batch.
%
%we draw 32 positive samples from each frame and 96 negative samples from each frame to form the input data in a mini-batch.
%
Note that we use RGBT234~\cite{li2019rgb} dataset as training dataset when conducting evaluation on GTOT~\cite{li2016learning} dataset.
In contrast, when performing evaluation on RGBT234 and RGBT210 datasets, we use  GTOT dataset as the training set as RGBT234 and RGBT210 have some overlaps in videos. 
%

%\section{Implementation Details}
\section{Online RGBT Tracking}
In the tracking process, we re-construct a new last fully connect layer as the video-specific layer for each instance object in each sequence.
Then, we freeze all the parameters of convolution layers ($W_{conv}$) and fine-tune the three fully connected layers ($W_{fc4,fc5,fc6}$) using the initial target state.
Specifically, given the first frame pair of the sequence and the ground truth bounding box, we draw 500 positive and 5000 negative samples as the training samples, where we define the samples whose IoU with the ground truth is larger than 0.7 as positive samples and smaller than 0.5 as negative samples. 
Note that we employ three different methods for sample generation, including Gaussian sampling, uniform sampling and global sampling~\cite{nam2016learning}.
In both training and inference phases, we use the Gaussian sampling to generate positive anchor boxes, and employ the uniform sampling and global sampling to generate negative anchor boxes.
In addition, in the inference phase, we also introduce the re-sampling method to generate more high-quality candidates for the compensation of sudden camera motion.
In the initial training process, we employ these samples to train the three fully connected layers with 50 iterations, and set the learning rate as 0.005 and 0.0005 for the last layer and other two fully connected layers, respectively.
We apply the bounding box regression technique~\cite{nam2016learning} to improve target localization accuracy and estimate the target scales during tracking process.
To prevent the potential unreliability, we only train a bounding box regressor in the initial frame, and use it to adjust target states in subsequent frames.
In subsequent frames, we use the Gaussian sampling to generate positive anchor boxes and uniform sampling to generate negative anchor boxes, and then collect 50 positive samples whose IoU with the ground truth box is larger than 0.7 and 200 negative samples whose IoU is smaller than 0.3 as training samples for short-term and long-term update~\cite{nam2016learning}.
The learning rates of the last fully connected layer and the other two fully connected layers are set to 0.01 and 0.001 respectively.
Given the $t$-th frame, we first draw a candidate set {$X_t^i$} from a Gaussian distribution of previous frame tracking result $X_{t-1}^*$, where the mean of Gaussian function is set to $X_{t-1}^* = (cx_{t-1}, cy_{t-1}, s_{t-1})$ and the covariance is set as a diagonal matrix $diag (0.09r^2,0.09r^2,0.25)$. 
where $r$ is the mean of the width and height of target, and the $(cx,cy)$ and $s$ indicate the location and scale respectively in the previous frame.
In tracking process, we feed all candidate samples from {$X_t^i$} into our network, and compute the positive scores and negative scores of candidate samples using the trained networks as $f^+(X_t^i)$ and $f^-(X_t^i)$, respectively.
We sort the candidate samples by their scores and select the candidate samples with the top five highest scores, and then compute its the mean value as the tracking result $X_t^*$ of the current frame $t$,
where the $X_m^*$ is denoted as the samples set with top five highest scores and the $mean()$ represents the averaging operation. 
The formula expression is as follows:
\begin{equation}
	\begin{aligned}
		&{\mathit{X}}_m^{\ast} = \mathop{ \arg\max}_{i=1,...,256}\ \mathrm{\mathit{f^+(X_t^i)}}\\
		&{\mathit{X}}_t^{\ast} = \mathop{ mean}(X_m^*).
		\label{eq::7} 
	\end{aligned}
\end{equation}
If the mean of the top five scores $F(X_t)$ are less than 0, we calculate the average moving vector $[dx,dy]$ and judge abrupt camera motion to be occurred when the amplitude of $[dx,dy]$ exceeds the predefined threshold $u$.
Empirically, we set $u$ to 5 in this work, and its setting is validated in Fig~\ref{fig::U_value}.
We can see that when $u$ is equal to 0, the performance is the lowest at this time.
This is because all tracking failures are caused by camera motion by default and more interference candidate samples will inevitably be brought, causing performance degradation.
We also can see that when $u$ is equal to 5, the performance is the highest, and then as the value of $u$ increases, the performance has a downward trend. This is because the larger the value of $u$, the fewer failure cases that are recognized as camera movement, and thus the resampling mechanism cannot be activated.
When the abrupt camera motion occurs, we execute the re-sampling strategy to obtain a new candidate set and then feed it into our network to compute their scores, obtaining the top score $RF(X_t)$.
To improve the robustness, we combine the Gaussian sampling and re-sampling methods to determine the final tracking results.
In specific, we use the candidate sample with higher score from $F(X_t)$ and $RF(X_t)$ as the predicted tracking result.
More details can be referred to Algorithm~\ref{alg::tracking_alg}.

\begin{algorithm}[htbp]  
	\caption{Online RGBT Tracking Process}  
	\label{alg::tracking_alg}  
	
	\begin{algorithmic}[1]  
		\Require   
		Pretrainded CNN filters {$W_{conv}$,$W_{fc4,fc5}$};
		Initial target state $X_1$; 
		Threshold $u$.
		\Ensure 
		
		Estimated target state $X^*_t$.
		
		% \bf{Initialize:}
		\State Randomly initialize the last layer $W_{fc6}$;
		
		\State Train a bounding box regression model $BB(\cdot)$~\cite{nam2016learning} ;
		
		\State Draw positive samples $S^+_1$ and negative samples $S^-_1$;
		
		\State Fine-tune $W_{fc4,fc5,fc6}$ using $S^+_1$ and $S^-_1$;
		
		\State Initialize short-term~\cite{nam2016learning} and long-term~\cite{nam2016learning} sample set $\phi_s,\phi_l$.
		
		\Repeat 
		\State Draw target candidate set $X_t^i$;  
		
		\State Compute target state $X_m^*$ and score $F(X_t)$ by Eq.~\ref{eq::5}.

		\If {$F(X_t) > 0$} 
		
		\State Estimated target state: $X_t^* = mean(X_m^*)$;
		
		\State $X_t^* = BB(X_t^*)$;
		
		\State Update {$\phi_s,\phi_l$}.
		
		\Else 
		
		\State Compute $[dx,dy]$ by Disflow~\cite{disflow};
		
		\If {$|dx| > u$ or $|dy| > u$}
		
		\State Perform re-sampling;
		
		\State Compute target state $RX_t^*$ and score $RF(X_t)$. 
		
		\If{$RF(X_t) > (F(X_t)$}
		
		\State Estimate target state: $X_t^* = RX_t^*$.
		
		\Else
		
		\State Estimate target state: $X_t^*$.
		
		\EndIf
		
		\Else
		
		\State Execute short-term update {$W_{fc4,fc5,fc6}$} using {$\phi_s$}.
		
		\EndIf
		
		\EndIf
		
		\If {$t$ mod 10 = 0}
		
		\State Execute long-term update {$W_{fc4,fc5,fc6}$} using {$\phi_l$}.
		
		\EndIf 
		
		\Until{end of sequence}  
	\end{algorithmic}  
\end{algorithm} 

\begin{table*}[t]
	\centering
	\setlength{\belowcaptionskip}{0.5cm}
	\caption{Attribute-based PR/SR scores (\%) of our DMCNet on RGBT234 dataset against with eight RGBT trackers. The best, second and third results are in {\color{red}red},{\color{blue}blue} and {\color{green}green}.}\smallskip
	\resizebox{1\textwidth}{!}{
		\begin{tabular}{ c | c  c  c  c  c  c  c  c  c c  |  c  }
			\hline
			Trackers & MDNet+RGBT &SGT & CMR & DAPNet & MANet & MaCNet & FANet & CMPP & JAMMC &CAT  & DMCNet \\
			\hline
			Pub. Info.& CVPR2016 &  ACM MM2017 & ECCV2018  & ACM MM2019 & ICCVW2019 & Sensors2020 & TIV2020 & CVPR2020 & TIP2020 &ECCV2020  &  \\
			\hline
			NO & 86.2/61.1 & 87.7/55.5  & 89.5/61.6  & 90.0/64.4  & 88.7/64.6  & 92.7/66.5  & 88.2/65.7  & {\color{red}95.6}/{\color{blue}67.8}  & {\color{blue}93.2}/{\color{red}69.4}  & {\color{blue}93.2}/66.8   &  {\color{green}92.3}/ {\color{green}67.1} \\
			
			PO & 76.1/51.8 & 77.9/51.3  & 77.7/53.6  & 82.1/57.4  & 81.6/56.6  & 81.1/57.2  & {\color{blue}86.6}/{\color{green}60.2}  & {\color{green}85.5}/60.1  & 84.1/{\color{blue}61.1} & 85.1/59.3  &  {\color{red}89.5/63.1} \\ 
			
			HO & 61.9/42.1 & 59.2/39.4  & 56.3/37.7  & 66.0/45.7  &68.9/46.5  & {\color{green}70.9/48.8}  & 66.5/45.8  & {\color{blue}73.2/50.3}  & 67.7/48.3 & 70.0/48.0  &  {\color{red}74.5/52.1}\\
			
			LI &67.0/45.5 & 70.5/46.2  & 74.2/49.8  &77.5/53.0  &76.9/51.3  & 77.7/52.7  & 80.3/54.8  & {\color{red}86.2}/{\color{green}58.4}  & {\color{green}84.0}/{\color{red}58.8} &81.0/54.7   & {\color{blue}85.3}/{\color{blue}58.7}\\
			
			LR &67.0/45.5  & 72.5/46.2  & 72.0/47.6  & 75.9/51.5  &70.8/48.7  & 75.1/47.6  & 75.0/51.0  & {\color{red}86.5}/{\color{green}57.1}  & {77.1}/{51.7} &{\color{green}82.0}/{\color{green}53.9}  & {\color{blue}85.4}/{\color{red}57.9}\\
			
			TC & 75.6/51.7 & 76.0/47.0  & 67.5/44.3  & 76.8/54.3  &75.4/54.3  &77.0/56.3  & 76.6/54.9  & {\color{blue}83.5/58.3}  & 74.9/52.6 & {\color{green}82.0/53.9} & {\color{red}87.2/ 61.2}\\
			
			DEF &66.9/47.3 & 68.5/47.4 & 66.7/47.3  & 71.7/57.8  &72.0/52.4  & 73.1/51.4  & {\color{green}72.2}/52.6  & 75.0/{\color{blue}54.1}  & 70.6/{\color{green}52.9} &{\color{blue}76.2/54.1} & {\color{red}77.9/56.5}\\
			
			FM &58.6/36.3 & 67.7/40.2  & 61.3/38.4  & 67.0/44.3  &69.4/44.9  & 69.4/44.9  & 68.1/43.6  & {\color{blue}78.6/50.8} & 61.0/41.7 & {\color{green}73.1/47.0}  & {\color{red}80.0/52.4}\\
			
			SV  &73.5/50.5 &69.0/43.4  & 71.0/49.3  & 78.0/54.2  &77.7/54.2  & 78.7/56.1  & 78.5/56.3  &  {\color{green}81.5/57.2} &  {\color{blue}83.7}/ {\color{red}61.6} &79.7/56.6  &  {\color{red}84.6}/ {\color{blue} 59.8}\\
			
			MB &65.4/46.3 & 64.7/43.6  & 60.0/42.7  & 65.3/46.7  &72.6/51.6  & 71.6/52.5  &70.0/50.3  & {\color{blue}75.4}/{\color{green}54.1} & {\color{green}75.1}/{\color{blue}54.9}  &68.3/49.0    & {\color{red}77.3/ 55.9}\\
			
			CM  &64.0/45.4 & 66.7/45.2  & 62.9/44.7  & 66.8/47.4  &71.9/50.8  & 71.7/51.7 & 72.4/52.3 & {\color{green}75.6/54.1} & {\color{blue}76.2/55.6} &75.2/52.7  &{\color{red}80.1/ 57.6}\\
			
			BC  &64.4/43.2 & 65.8/41.8  & 63.1/39.8  & 71.7/48.4  &73.9/48.6  & 77.8/50.1  & 75.7/50.2  & {\color{blue}83.2/53.8 } & 68.7/48.5 &{\color{green} 81.1/51.9}  &{\color{red}83.8/ 55.9}\\
			\hline
			ALL &72.2/49.5 &72.0/47.2  & 71.1/48.6  & 76.6/53.7  &77.7/53.9  & 79.0/55.4   & 78.7/55.3  & {\color{blue}82.3/57.5} & 79.0/{\color{green} 57.3} & {\color{green} 80.4}/56.1  & {\color{red}83.9/ 59.3} \\
			\hline
	\end{tabular}}
	\label{tb::AttributeResults}
\end{table*}

\section{Performance Evaluation}

In this section, we evaluate our duality-gated mutual-conditional network (named DMCNet in this paper) with existing RGBT and RGB trackers on four popular RGBT tracking benchmark datasets including GTOT~\cite{li2016learning}, RGBT210~\cite{Li17rgbt210}, RGBT234~\cite{li2019rgb} and VOT-RGBT2019~\cite{vot-rgbt2019}.
The experimental environment is configured as follows: Pytorch 1.0+, 8 NVIDIA GeForce GTX 2080Ti GPU server.

\subsection{Evaluation Setting}

{\flushleft \bf Datasets}.
We use four large challenging tracking datasets, including GTOT and RGBT210, RGBT234 and VOT-RGBT2019, to comprehensively evaluate our DMCNet. 
GTOT consists of 50 aligned RGB and thermal infrared video pairs, containing approximately 15K frames of images and seven visual tracking challenge attributes. 
RGBT210 consists of 210 aligned RGB and thermal infrared video pairs, containing 210K frames in total and a maximum of 8K frames per video pair and a total of 12 visual tracking challenge attributes. 
RGBT234 dataset, an extension of the RGBT210~\cite{Li17rgbt210} tracking dataset, consists of 234 aligned RGB and thermal infrared video pairs, containing approximately 200K frames of images and 12 visual tracking challenge attributes, such as camera moving, large scale variations and environmental challenges.
VOT-RGBT2019 dataset, a subset of the RGBT234 dataset~\cite{li2019rgb}, which consists of 60 high aligned RGBT video sequences selected from RGBT234 dataset~\cite{li2019rgb}, with a total of over 20K frames. 

{\flushleft \bf Evaluation metrics}.
Precision rate (PR) and success rate (SR) are used to evaluate RGBT tracking performance on three RGBT tracking datasets.
PR is the percentage of frames whose distance of the output position with the ground truth is below a predefined threshold, and the thresholds in the GTOT and RGBT234 tracking datasets are set to 5 and 20 pixels respectively to obtain the representative PR score (because the target objects in the GTOT dataset are usually small). 
SR is the percentage of frames where the overlap rate between the output bounding boxes and the ground truth bounding boxes are greater than a threshold.
By changing the threshold, SR curve can be obtained, and the area under the curve of SR curve is used to define the representative SR.
To more comprehensive evaluate different RGBT tracking algorithms, we also follow the VOT official evaluation protocol.
Specifically, here three evaluation metrics, Expected Average Overlap (EAO), robustness (R) and accuracy (A), are used.
A is the average overlap between the predicted and ground truth bounding boxes during successful tracking periods.
R measures how many times the tracker loses the target (fails) during tracking.
EAO is a combined measure of A and R.

\subsection{Evaluation on GTOT Dataset}
\begin{figure}[t]
	\centering
	\includegraphics[width=0.49\textwidth]{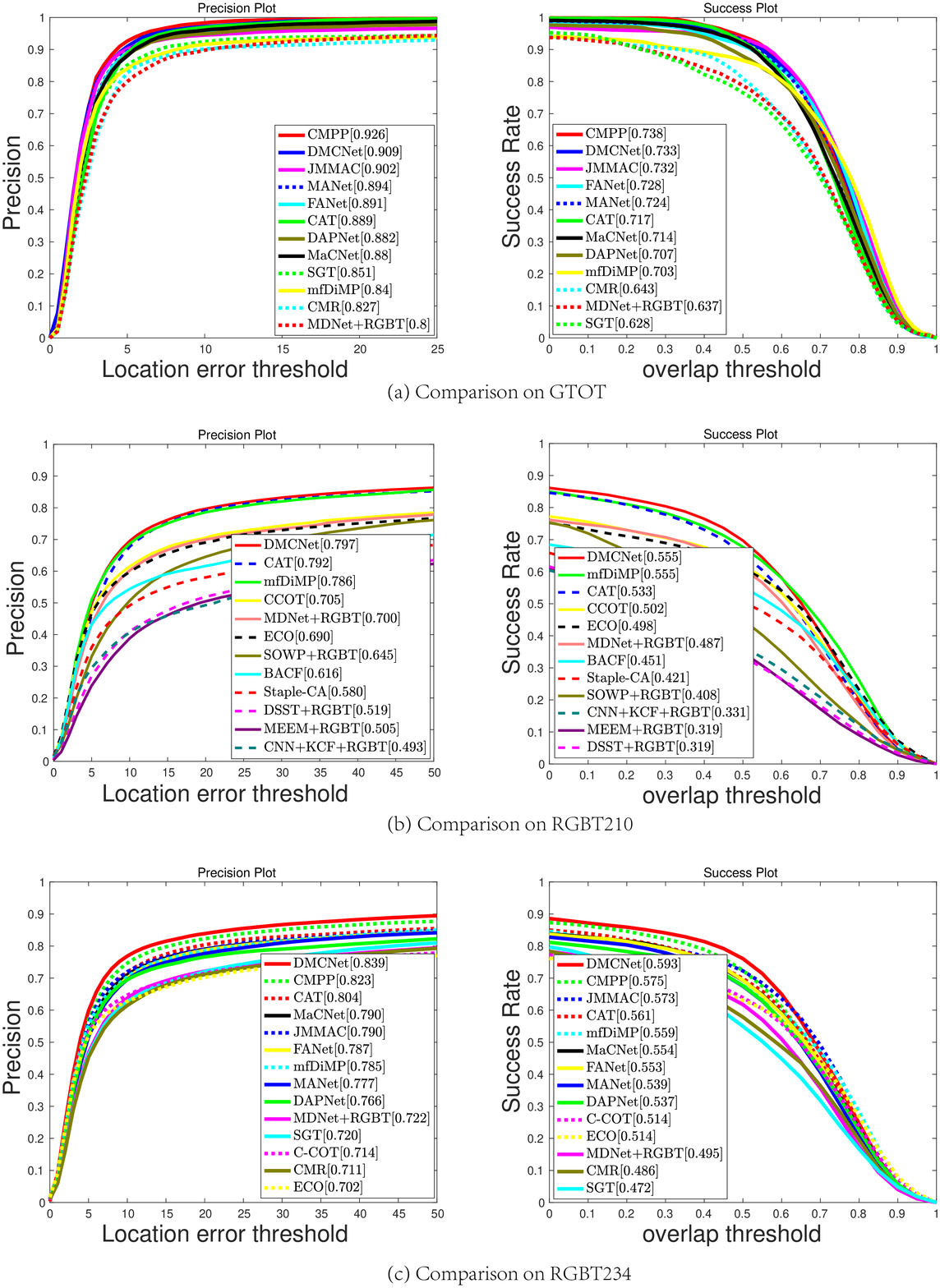} 
	\caption{Evaluation curves on GTOT, RGBT210 and RGBT234 datasets compare with RGBT trackers. The representative scores of PR/SR are presented in the legend.}
	\label{fig::result_210_234}
\end{figure}

In GTOT dataset, we compare our DMCNet with 11 state-of-the-art RGBT trackers, among which SGT~\cite{Li17rgbt210}, DAPNet~\cite{zhu2019dense}, MANet~\cite{li2019manet}, MaCNet~\cite{zhang2020MaCNet} and CMR~\cite{Li18eccv}, CMPP~\cite{2020CMPP}, CAT~\cite{2020CAT}, FANet~\cite{2020FANet}, mfDiMP~\cite{zhang2019mfdimp}, JMMAC~\cite{2020JMMAC} are RGBT trackers, while MDNet~\cite{nam2016learning} +RGBT are the extended RGBT tracker from existing RGB tracking algorithm by concatenating thermal and RGB features.
%while Struck~\cite{hare2015struck} +RGBT, DAT~\cite{Pu2018Deep} +RGBT, MDNet~\cite{nam2016learning} +RGBT, RT-MDNet~\cite{RT-MDNet18eccv} +RGBT, SiamDW~\cite{zhang2019deeper} +RGBT are the extended RGBT trackers from existing RGB tracking algorithms by regarding the thermal as an additional channel or concatenating thermal and RGB features.
%
The evaluation results are shown in Fig.~\ref{fig::result_210_234} (a), and our DMCNet has a comparable performance with the state-of-the-arts on GTOT dataset.
%
%Compare with CMPP~\cite{2020CMPP} method, which employs a lot of history samples as temporal information to enhance the current frame representation, while our DMCNet just use the current frame information.
%
%
It is worth noting that GTOT dataset contains 50 short video sequences (about 40-250 frames per sequence) and is thus less challenging than RGBT234 dataset.
Therefore, there are two reasons why CMPP gets higher performance than DMCNet. 
On one hand, CMPP constructs a historical information pool using external storage, which is more effective in tracking short video sequences as the memory pool is not easily contaminated. 
On the other hand, GTOT includes many small tracking objects, and CMPP applies the feature pyramid method to aggregate the features of all layers to enhance the feature representation ability of small objects. 
Although CMPP has achieved state-of-the-art performance on GTOT, our algorithm achieves comparable performance in SR score and better tracking efficiency. 
On more challenging RGBT234 dataset, our method outperforms CMPP clearly. 
These results fully demonstrate the effectiveness and efficiency of our method against CMPP.
%
%Note that, due to GTOT dataset no camera motion scenes, thus the RS module is not executed in GTOT evaluation.
%
%Compared with other methods, it outperforms the second best tracker MANet~\cite{li2019manet} with 1.1\%/0.2\% gains in PR/SR.
%
%Overall, the excellent performance demonstrates the effectiveness of our proposed method.
%

\subsection{Evaluation on RGBT210 Dataset}
In RGBT210~\cite{Li17rgbt210} dataset, we compare our method with 11 trackers using two evaluation metrics. 
From Fig.~\ref{fig::result_210_234} (b), we can see that the performance of our method exceeds mfDiMP~\cite{zhang2019mfdimp} 1.1\% in PR and the CAT~\cite{2020CAT} 2.2\% in SR , and significantly outperforms other trackers, including CCOT~\cite{C-COT16eccv}, MDNet~\cite{nam2016learning} +RGBT, ECO~\cite{ECO17cvpr}, SGT~\cite{Li17rgbt210}, SOWP~\cite{kim2015sowp} +RGBT, DSST~\cite{martin2014DSST} +RGBT, BACF~\cite{kiani2017BACF}, Staple-CA~\cite{bertinetto2016staple}, MEEM~\cite{MEEM14eccv} +RGBT and CNN +KCF~\cite{2015KCF} +RGBT
 
It is worth noting that mfDiMP~\cite{zhang2019mfdimp} is the winner of the VOT2019-RGBT tracking competition.
Furthermore, it uses a larger-scale generated RGBT dataset (9335 sequences) to train network, but our network only uses GTOT dataset (50 sequences) as training set.

\begin{figure*}[t]
	\centering
	\includegraphics[width=0.95\textwidth]{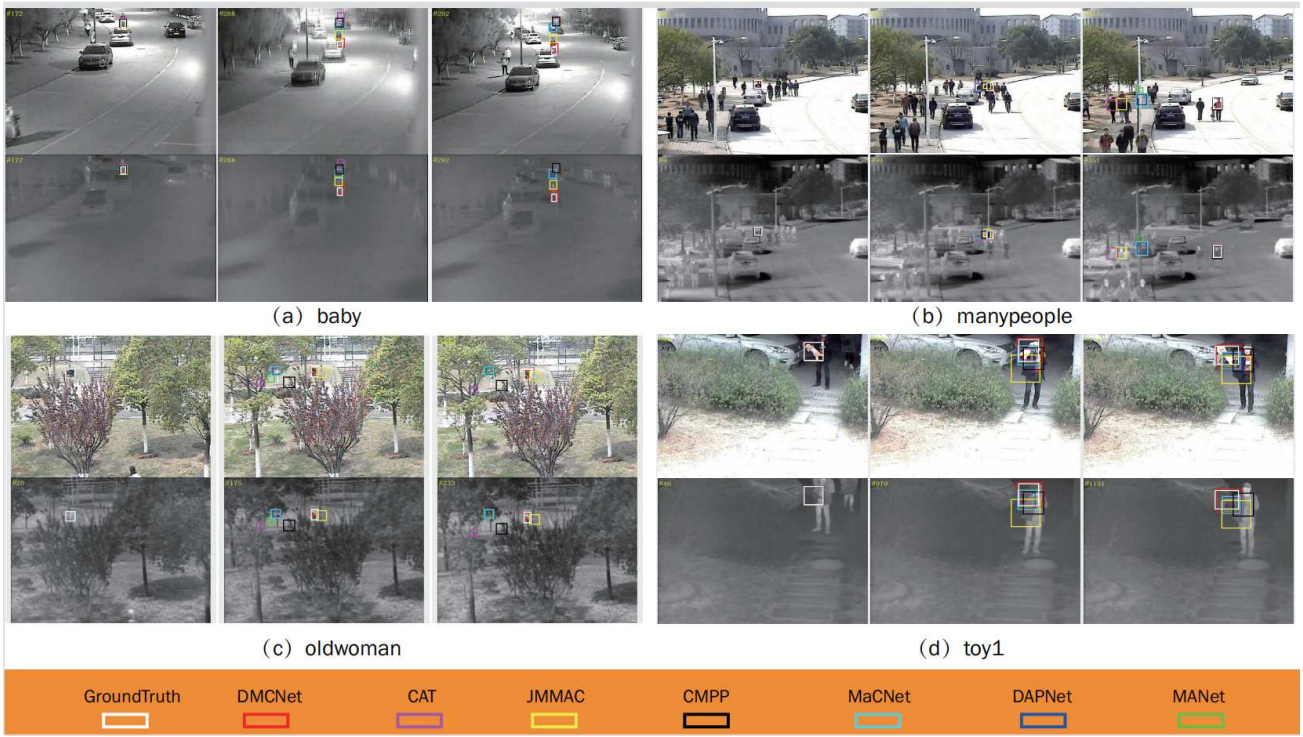}
	\caption{ Qualitative comparison of DMCNet against six state-of-the-art trackers on four video sequences. (a) The $baby$ sequence with the challenges of camera moving and low resolution and low illumination, (b) the $manypeople$ sequence with background clutter and heavy occlusion, (c) the $oldwoman$ sequence with low resolution and camera moving, and (d) the $toy1$ sequence with thermal crossover. For each sequence, the top row shows the frames of RGB modality while the bottom one shows the frames of thermal modality. For the results of different trackers, we use different color rectangles to represent them and the details are presented in the legend.}
	\label{fig::visual_results}
\end{figure*}

\subsection{Evaluation on RGBT234 Dataset}

To further evaluate our method, we conduct the experiments on RGBT234 tracking dataset, including overall comparison and challenge-based performance.

{\flushleft \bf Overall comparison}.
For more comprehensive evaluation, we compare our DMCNet with 13 state-of-the-art trackers, among which CMPP~\cite{2020CMPP}, CAT~\cite{2020CAT}, JMMAC~\cite{2020JMMAC}, FANet~\cite{2020FANet}, mfDiMP~\cite{zhang2019mfdimp}, SGT~\cite{Li17rgbt210}, CMR~\cite{Li18eccv}, CCOT~\cite{C-COT16eccv}, ECO~\cite{ECO17cvpr}, MDNet~\cite{nam2016learning} +RGBT, DAPNet~\cite{zhu2019dense}, MANet~\cite{li2019manet} and MaCNet~\cite{zhang2020MaCNet} are RGBT trackers.
The evaluation results are shown in Fig.~\ref{fig::result_210_234} (c).
We can see that the performance of our DMCNet has a clearly superior comparing with the state-of-the-art RGBT methods in all metrics.
It fully demonstrates the effectiveness of our method. 
In particular, our DMCNet (83.9\%/59.3\% in PR/SR) achieves 1.6\%/1.8\%, 3.5\%/3.2\% and 4.9\%/2.0\% gains in PR/SR over CMPP~\cite{2020CMPP}, CAT~\cite{2020CAT} and JMMAC~\cite{2020JMMAC} respectively. 
Compared with MDNet-based RGBT tracker CAT, CMPP improves PR score by 1.9\% and SR score by 1.4\%, but our algorithm achieves a great increase of 3.5\%/3.2\% in PR/SR.
It is worth noting that CMPP needs to construct a historical information pool, which not only requires external storage memory, but also increases the computational burden.
However, our algorithm achieves better performance and efficiency without using external storage.

%%%%%%%%%%%%%%%%%%%%%%%%%%%暂定%%%%%%%
%{\color {red}
%The main reason is our algorithm more robust than mfDiMP. Compared with VOT-RGBT2019 dataset containing only 60 sequences, the RGBT234 dataset has a total of 234 sequences, which contain more challenging factors. Therefore, our method is more robust than mfDiMP and achieves better performance on RGBT234 dataset.
%}

In addition, to further validate the effectiveness of DMCNet, 
we take some advanced trackers to compare in Fig.~\ref{fig::visual_results}, including CMPP~\cite{2020CMPP}, CAT~\cite{2020CAT}, JMMAC~\cite{2020JMMAC}, MANet~\cite{li2019manet}, DAPNet~\cite{zhu2019dense} and MaCNet~\cite{zhang2020MaCNet}.
Specifically, we present some visual tracking results on four sequences, and for each sequence we presents three frame pairs for the clarity. 
From Fig.~\ref{fig::visual_results}, we can see that our approach performs obviously better than other trackers in several challenges, such as camera motion, thermal crossover, background clutter, heavy occlusion, low resolution and low illumination.
For instance, Fig.~\ref{fig::visual_results} (a) and (c) show the tracking results of our method with other trackers on the video sequences with camera moving, low resolution and low illumination. 
Obviously, DMCNet can more robustly localize the target while other algorithms lose the tracked object when the sudden camera motion happens.
In Fig.~\ref{fig::visual_results} (b), the sequence has the challenges of background clutter and heavy occlusion, and most trackers are failed. While our method and CMPP~\cite{2020CMPP} can continuously track the target.
The video sequence shown in Fig.~\ref {fig::visual_results} (d) has a serious thermal crossover phenomenon. 
Other trackers just locate the target in part of frames, but our tracker can deal with this challenge well. 
To sum up, DMCNet is very robust in adverse conditions due to the benefits from the duality-gated mutual condition module and the re-sampling strategy.

\begin{table*}[!h]
	\caption{Comparison results on VOT-RGBT2019 dataset.}
	\label{tb::VOT_result}
	\centering
	\footnotesize %
	\begin{tabular}{c| c c c c c c c |c }
		\toprule %
		&GESBTT~\cite{vot-rgbt2019} &CISRDCF~\cite{vot-rgbt2019} &MPAT~\cite{vot-rgbt2019} &FSRPN~\cite{vot-rgbt2019} &mfDiMP~\cite{zhang2019mfdimp} & MaCNet~\cite{zhang2020MaCNet} & MANet~\cite{li2019manet} & DMCNet   \\
		\midrule
		A($\uparrow$)&0.6163&0.5215&0.5723 &0.6362 &0.6019 &0.5451  &0.5823 &0.6009  \\
		R($\uparrow$)&0.6350&0.6904&0.7242 &0.7069 &0.8036 &0.5914  &0.7010 &0.7088 \\
		\midrule
		EAO &0.2896&0.2923&0.3180 &0.3553 &0.3879 &0.3052  &0.3463  &0.3796 \\
		%\bottomrule
		\hline
	\end{tabular}
\end{table*}

{\flushleft \bf Challenge-based performance}.
We show the results of our DMCNet against other state-of-the-art RGBT trackers, including MDNet~\cite{nam2016learning} +RGBT, SGT~\cite{Li17rgbt210}, DAPNet~\cite{zhu2019dense}, MANet~\cite{li2019manet}, MaCNet~\cite{zhang2020MaCNet}, CMPP~\cite{2020CMPP}, CAT~\cite{2020CAT}, FANet~\cite{2020FANet} and JMMAC~\cite{2020JMMAC} on different subsets with different challenge attributes.
The challenge attributes include no occlusion (NO), partial occlusion (PO), heavy occlusion (HO), low illumination (LI), low resolution (LR), thermal crossover (TC), deformation (DEF), fast motion (FM), scale variation (SV), motion blur (MB), camera moving (CM) and background clutter (BC).  
The evaluation results are shown in Table~\ref{tb::AttributeResults}.
The results show that the our method performs the best under the most challenging conditions.
It demonstrates the robustness of our DMCNet in handling most adverse conditions. 

\subsection{Evaluation on VOT-RGBT2019 Dataset}
We further evaluate our method against several state-of-the-art trackers on VOT-RGBT2019~\cite{vot-rgbt2019} dataset, including mfDiMP~\cite{zhang2019mfdimp}, MaCNet~\cite{zhang2020MaCNet}, MANet~\cite{li2019manet}.
It is worth noting that GESBTT, CISRDCF, MPAT and FSRPN are the participating algorithms of the VOT-RGBT2019 competition. 
So far, none of these algorithms has been published in formal literature.
Therefore, we cite the official document of VOT-RGBT2019, which contains the performance of all participating algorithms.
We follow the VOT protocol and adopt EAO, R and A as the metrics.
We compare DMCNet with seven RGBT tracking algorithms and directly use the results reported in papers to ensure the best performance.
From the results of Table~\ref{tb::VOT_result}, we can see that our DMCNet has comparable performance against mfDiMP, and outperforms other state-of-the-art methods including MANet and MaCNet.
Some state-of-the-art RGBT tracking methods, such as CMPP, CAT, do not report the VOT2019-RGBT evaluation results, thus we do not consider the comparison with these algorithms in this dataset.
Note that, from the results we can find that our DMCNet lower than mfDiMP in R metric.
It is mainly due to two reasons. First, mfDiMP employs the conception of IoU network to improve tracking results. 
Second, mfDiMP uses the training set of GOT-10k dataset to generate a large-scale synthetic RGBT dataset as their training data (9,335 videos with 1,403,359 frames in total), while we only use GTOT dataset (50 videos with 15,000 frames in total) to train our network. 
We will improve the performance of DMCNet from these considerations in the future.
Although the performance of DMCNet is lower than mfDimp on VOT2019-RGBT dataset, it outperforms mfDiMP on RGBT234 dataset.
The main reason is our algorithm more robust than mfDiMP. 
Compared with VOT-RGBT2019 dataset containing only 60 sequences, RGBT234 dataset has a total of 234 sequences, which contains more challenging factors. 
Therefore, our method is more robust than mfDiMP and achieves better performance on RGBT234 dataset.
Although FSRPN~\cite{vot-rgbt2019} has higher and closer performance than DMCNet in A and R metrics, we exceed this method by $2.43\%$ in EAO metric.
The reason is that the value of EAO is not only computed by the values of A and R. 
In the VOT protocol, if a tracker fails in a sequence, it will be reinitialized in the five frames later. 
Thus R only reflects the number of failures, while the EAO is also related to the location of failures. 
Even though two trackers have the similar R values, the different failure positions in the sequence will result in different EAO values. 
Therefore, the reason why EAO of FSRPN is lower than DMCNet is that the robustness of FSRPN is lower and fails earlier than DMCNet.

\subsection{Analysis of Our DMCNet}

\begin{figure}[htbp]
	\centering
	\includegraphics[width=0.45\textwidth]{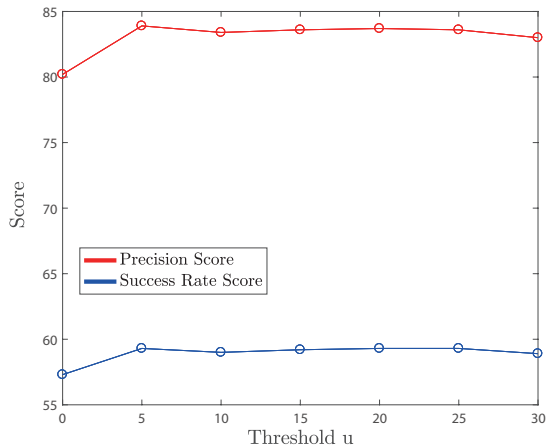} % Reduce the figure size so that it is slightly narrower than the column.
	\caption{Results of our method with different $u$ values on RGBT234 dataset.
	}
	\label{fig::U_value}
\end{figure}

{\flushleft \bf Impact of parameter $u$}.
In the module of re-sampling, $u$ is a critical hyper parameter. 
We manually set $u$ =\{ 0, 5, 10, 15, 20, 25, 30\} to analyze the impact on the tracking performance.
The tracking results on RGBT234 dataset are shown in the Fig.~\ref{fig::U_value}.
From the results, we can see the tracker achieves the best performance with $u$=5.
By observing Fig.~\ref{fig::U_value}, we can find that the tracking performance first rises and then falls with the increasing of $u$ value.
The main reason is that the re-sampling module will more easily be executed when $u$ is smaller, which will cause more false activations.

{\flushleft \bf Benefits from low-quality data}.
To verify our core idea that low-quality modalities benefit the results, we conduct experiments as follows.
We remove low-quality modalities in all datasets, including the thermal modality sequences with thermal crossover and the RGB modality sequences with low illumination, denoted as GTOT-V1 and RGBT234-V1 datasets. 
In this end, RGBT234-V1 contains 227 video sequences, and 77 of them contain only one modality.
Similarly, GTOT-V1 contains 36 video sequences, and 30 of them contain only one modality.
For sequences of single modality, we take same modality as inputs in our network.
We evaluate our method on these datasets, and the results are shown in Table~\ref{table::low_data_table}.
We can see that after removing the low-quality data, the performance of our method has dropped in both GTOT-V1 and RGBT234-V1 datasets. 
In other words, if low-quality data are introduced, our algorithm can well leverage their effective features for performance boosting.
Note that when we directly discard low-quality data in tracking, the performance drops significantly. 
It is because low-quality data contain much useful information benefiting for tracking in spite of the presence of many noises. 
Our method can effectively learn features from all modalities even in the presence of low-quality modalities.

\begin{figure*}[ht]
	\centering
	\includegraphics[width=0.95\textwidth]{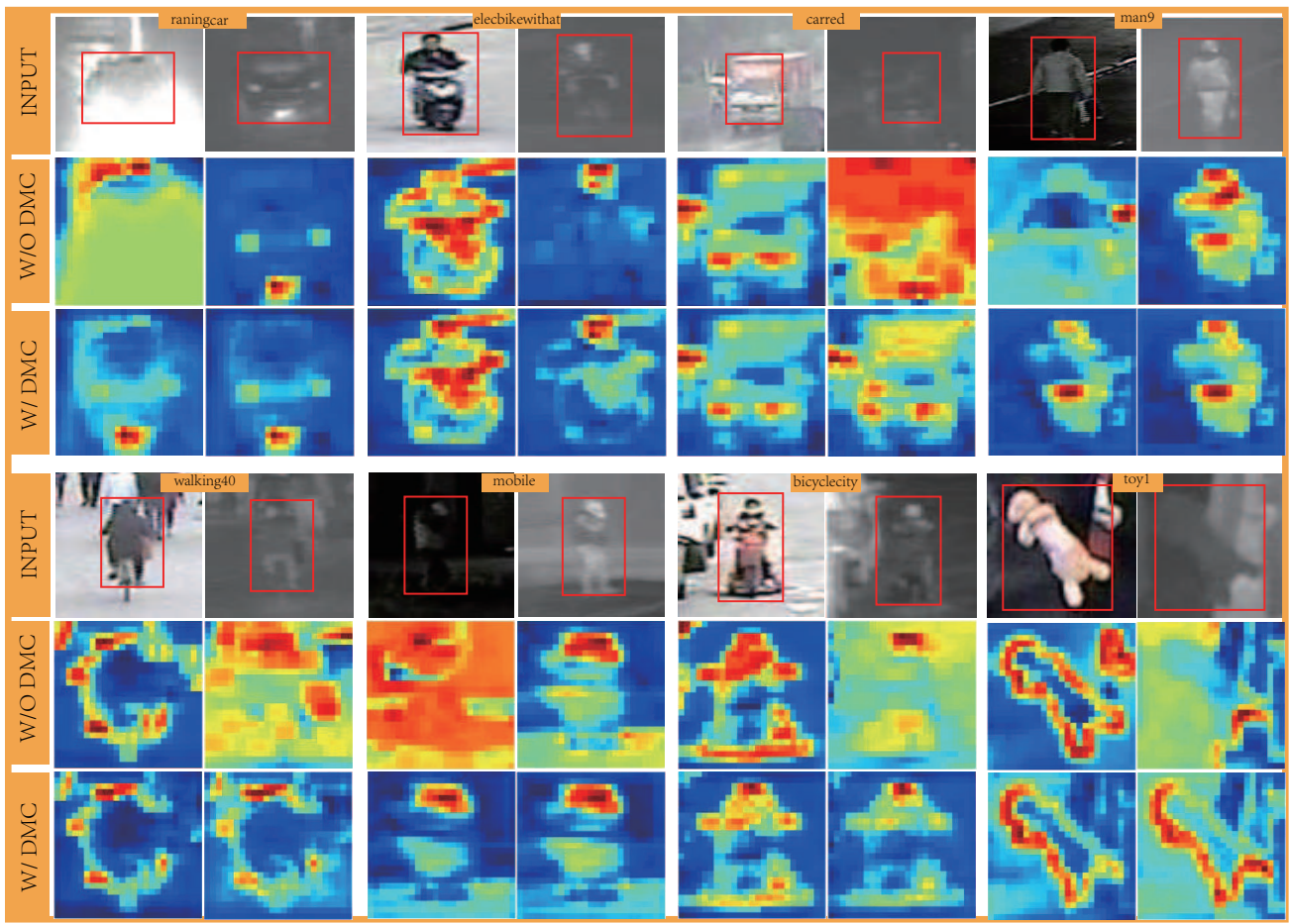}
	\caption{Qualitative comparison between w/o DMC and w/ DMC on eight RGBT video sequences.}
	\label{fig::visual_exp_featuremap}
\end{figure*}

To further validate the effectiveness of the DMC module in data noise suppression in low-quality modality, we present the visualization with w/ DMC and w/o DMC of our method in Fig.~\ref{fig::visual_exp_featuremap}. Herein, w/ denotes the ``with'' operation and w/o denotes the ``without'' operation. 
Here the first and fourth row are inputs of our network, the second and fifth row are output feature maps of the modality-specific sub-network without DMC module, and the third and sixth row are output feature maps of modality-specific sub-network with DMC module.
From Fig.~\ref{fig::visual_exp_featuremap} we can see that w/ DMC can better suppress the noises than w/o DMC in low-quality modality.
For example, in the $mobile$ and $toy1$ sequences, both modalities contain many noises, but our DMC module can suppress most of noises well by leveraging effective features of all modalities.
In other examples, even if one modality is low-quality, our method can also suppress data noises with the help of high-quality modality.

\setlength{\tabcolsep}{3pt}
\begin{table}[t]
	\begin{center}
		\caption{ PR/SR scores of our methods on RGBT234 and GTOT datasets and their two sub-datasets.}
		\begin{tabular}{ccc|ccc}
			\hline
			DMCNet & PR   & SR    &DMCNet & PR   & SR           \\
			\hline
			GTOT    & \textbf{0.909}&\textbf{0.733}  &RGBT234  & \textbf{0.839}&\textbf{0.593} \\
%			GTOT-V1 & 0.898  & 0.723    &RGBT234-V1 &0.829  & 0.592  \\
			GTOT-V1 & 0.782    &0.692    &RGBT234-V1 & 0.820    &0.584  \\
			\hline
		\end{tabular}
		\label{table::low_data_table}
	\end{center}
\end{table}
\setlength{\tabcolsep}{1pt}

{\bf Benefits from complementary multiple modalities}.
To validate the effectiveness of multi-modal data, we design three variants and evaluate them on RGBT234 and GTOT datasets.
There are:
1) DMCNet-AVG-RGBT, which uses the average image of two modality images as input of our network.
2) DMCNet-RGB, which only uses the RGB modality image as input of our network. 
3) DMCNet-T, which only uses the thermal modality image as input of our network.
Table~\ref{table::ablation_table_1} presents the comparison results on RGBT234 and GTOT datasets, and the results fully demonstrate the effectiveness of the multiple modalities and adaptive fusion of different modalities in our method.
Specifically, the performance of DMCNet-AVG-RGBT is better than DMCNet-RGB and DMCNet-T on RGBT234 dataset, which suggests that the complementary advantages of RGB and thermal modalities can boost tracking performance.
In GTOT dataset, the performance of DMCNet-AVG-RGBT is lower than DMCNet-RGB, which indicates that the direct fusion of the two modalities could not avoid noises from weak modalities.
While our DMCNet achieves great performance improvement over all of them on both datasets. It shows the effectiveness of our proposed method in the adaptive fusion of different modalities.

\setlength{\tabcolsep}{3pt}
\begin{table}[t]
	\begin{center}
		\caption{PR/SR scores of different inputs in our method on RGBT234 and GTOT datasets. }
		\begin{tabular}{cccccccc}
			\hline
			&     &    & \multicolumn{2}{c}{RGBT234} & \multicolumn{2}{c}{GTOT} \\
			\hline
			&     &			 & PR           & SR           & PR          & SR       \\
			\hline
			DMCNet & & & \textbf{0.839} &\textbf{0.593} &\textbf{0.909} & \textbf{0.733}         \\			
			DMCNet-AVG-RGBT &     &    & 0.761        & 0.538  &0.779&0.636  \\
			DMCNet-RGB   &     &  & 0.740        & 0.526  &0.800&0.664 \\
			DMCNet-T &     & &0.703 &0.482 &0.760&0.624  \\
			\hline
		\end{tabular}
		\label{table::ablation_table_1}
	\end{center}
\end{table}
\setlength{\tabcolsep}{1pt}

{\flushleft \bf Ablation study}.
To validate the effectiveness of major components in our method, we implement three variants and evaluate them on RGBT234 and GTOT datasets.
They are:
1) DMCNet-v1, that removes all duality gated mutually conditional modules and the resampling strategy in our DMCNet;
2) DMCNet-v2, that removes the resampling strategy in our DMCNet;
3) DMCNet-v3, that removes all duality-gated mutual condition modules in our DMCNet.
Table~\ref{table::ablation_table} presents the comparison results on RGBT234 and GTOT datasets, and the results demonstrate the effectiveness of the proposed components. 
It is worth noting that the results of DMCNet-v3 on RGBT234 dataset show that our DMC module can learn more discriminative features in the challenges of motion blur caused by camera motion.
Moreover, we can see that the DMC module has the greatest impact on the speed of our DMCNet, and the RS module has a little impact on time cost.
The reason is that the RS module avoids many tracking failures, thereby greatly reduces the time cost on short-term updates.
Note that the RS module looks like useless on GTOT dataset, and the reason is that GTOT dataset does not include camera motion, i.e., the imaging cameras are fixed.
Therefore, our RS module will not be activated on GTOT dataset.

\setlength{\tabcolsep}{3pt}
\begin{table}[t]
	\begin{center}
		\caption{PR/SR scores of different variants induced from our method on RGBT234 and GTOT datasets. \Checkmark means adding the corresponding component.}
		\begin{tabular}{cccccccc}
			\hline
			&     &    & \multicolumn{2}{c}{RGBT234} & \multicolumn{2}{c}{GTOT} \\
			\hline
			& DMC & RS & PR           & SR           & PR          & SR    &FPS   \\
			\hline
			DMCNet-v1 & & & 0.802        & 0.565        & 0.866       & 0.693   &3.17      \\			
			DMCNet-v2 & \Checkmark &      & 0.820        & 0.584  &\textbf{0.909}&\textbf{0.733} &2.38 \\
			DMCNet-v3 &  & \Checkmark     & 0.809        & 0.569  &0.866&0.693 &3.16 \\
			DMCNet &\Checkmark&\Checkmark&\textbf{0.839}&\textbf{0.593}&\textbf{0.909}&\textbf{0.733} &2.38 \\
			\hline
		\end{tabular}
		\label{table::ablation_table}
	\end{center}
\end{table}
\setlength{\tabcolsep}{1pt}

%{\flushleft \bf DMC module ablation study.}
To validate the effectiveness of major components in our duality-gated mutual condition module, we implement five variants and evaluate them on RGBT234 and GTOT datasets.
They are:
1) DMC-w/o-msconv, that removes multi-scale convolutions in DMC;
2) DMC-w/o-gate, that removes all gates in DMC;
3) DMC-w/-one-gate, that removes all second gates in DMC.
4) DMC-w/o-shifting, that removes all shifting operations in DMC.
5) FiLM~\cite{perez2018film}, that replaces all DMC modules with the FiLM module.
Table~\ref{table::ablation_table2} presents the comparison results on RGBT234 and GTOT datasets, and the results demonstrate the effectiveness of these components in DMC. 
For the speed metric, we can see that the $msconv$ in our DMC module has a great influence, but other components have less influence in the DMC module.
From the above results, we can conclude that duality-gated strategy is truly bring more performance gains and each gate play a certain role in the DMC module.

%
%From the above results, we can conclude that the shifting operation acting on scaling later feature is better than original feature.
%
%Based on the above results, we can conclude that the shift operation applied to the scaled feature is better than that applied to the original feature.

\setlength{\tabcolsep}{3pt}
\begin{table}[t]
	\begin{center}
		\caption{ PR/SR scores of different DMC variants induced from our method on RGBT234 and GTOT datasets.}
		\begin{tabular}{cccccccccc}
			\hline
			&  \multicolumn{2}{c}{RGBT234} & \multicolumn{2}{c}{GTOT} \\
			\hline
			& PR   & SR         & PR   & SR     &FPS    \\
			\hline
			DMC    & \textbf{0.820}&\textbf{0.584} &  \textbf{0.909}&\textbf{0.733} &2.38 \\
			\hline
			DMC-w/o-msconv & 0.811     &0.575 & 0.901     &0.725  &2.85 \\

			DMC-w/o-gate    & 0.807  & 0.572   & 0.897  & 0.722 &2.28  \\
			
			DMC-w/-one-gate   & 0.813  & 0.577    &0.902   & 0.724  &2.43  \\
			
			DMC-w/o-shifting  & 0.790   &0.563    &89.9   &73.0  &2.32  \\
			
			FiLM~\cite{perez2018film}   & 0.812  & 0.565    &0.874   & 0.707 &2.64   \\
			\hline
		\end{tabular}
		\label{table::ablation_table2}
	\end{center}
\end{table}
\setlength{\tabcolsep}{1pt}

\section{Conclusion}
In this paper, we propose a duality-gated mutual condition network to make full use of the discriminative information of all modalities especially for low-quality modalities.
Our method employs mutual condition module to transform the effective information of RGB and thermal modalities as the mutual conditions, and then use them to fully enhance the discriminative ability of two modalities.
A duality-gated mechanism is also introduced to improve the quality of generated conditions.
Extensive experiments on four RGBT benchmark datasets show that our method achieves outstanding performance comparing with the state-of-the-art methods.
In future work, we will explore effective external knowledge to enlarge the power of duality-gated mutual conditions for more robust RGBT tracking.

\bibliographystyle{IEEEtran}
\bibliography{reference}

\end{document}